# Content-decoupled Contrastive Learning-based Implicit Degradation Modeling for Blind Image Super-Resolution

Jiang Yuan, Ji Ma, Bo Wang, Weiming Hu

*Abstract*—Implicit degradation modeling-based blind super-resolution (SR) has attracted more increasing attention in the community due to its excellent generalization to complex degradation scenarios and wide application range. How to extract more discriminative degradation representations and fully adapt them to specific image features is the key to this task. In this paper, we propose a new Content-decoupled Contrastive Learning-based blind image super-resolution (CdCL) framework following the typical blind SR pipeline. This framework introduces negative-free contrastive learning technique for the first time to model the implicit degradation representation, in which a new cyclic shift sampling strategy is designed to ensure decoupling between content features and degradation features from the data perspective, thereby improving the purity and discriminability of the learned implicit degradation space. In addition, we propose a detail-aware implicit degradation adapting module that can better adapt degradation representations to specific LR features by enhancing the basic adaptation unit's perception of image details, significantly reducing the overall SR model complexity. Extensive experiments on synthetic and real data show that our method achieves highly competitive quantitative and qualitative results in various degradation settings while obviously reducing parameters and computational costs, validating the feasibility of designing practical and lightweight blind SR tools.

## I. Introduction

Image super-resolution (SR) is a classic low-level vision task that aims to recover a low-resolution (LR) image into its high-resolution (HR) counterpart, which has been widely used in remote sensing, video surveillance, and medical imaging, etc. In previous studies, most SR methods [1]–[8] assume that the degradation process of the image is known and fixed (e.g., bicubic downsampling), which is called the non-blind SR. Although somewhat successful, non-blind SR methods do not fit into real-world scenarios, where the image degradation is usually complex and unknown. Once the degradation assumptions of the SR methods do not align with the actual degradation, the performance will be significantly reduced [9]. Therefore, how to realize high-quality image super-resolution with unknown degradation, called blind SR, has gradually become an emerging research topic [9]–[13].

The key to blind SR is to effectively estimate the degradation information in LR images. Early blind SR methods adopt an explicit degradation estimation strategy, which trains a degradation estimator to compute explicit degradation parameters from the LR image, e.g., the specific blur kernel or the noise level, as a guidance for the SR network. However, the application scope of explicit estimation-based blind SR is still limited. The unavoidable noise interference and the complex degradation combinations in real scenarios make it difficult to construct an accurate mathematical model of the degradation process, resulting in the method based on explicit estimation can only deal with some simple, low-order degradation problems. In addition, accurate numerical estimation often requires more parameters and iterations [9], making these methods time-consuming. To solve above problems, the blind SR community is gradually converting to the implicit degradation estimation strategy, which is the focus of this paper. The implicit strategy considers the degradation process to be fuzzy, so it models the degradation process as an indirect representation in the latent space. This latent representation is in line with the human fuzzy way of thinking [14], and can model the degradation process more comprehensively, which makes the estimator have stronger generalization and better robustness. At the same time, implicit degradation estimation strategy does not require extensive iterative optimization, thus significantly reducing computational costs.

The mainstream implicit estimation-based approaches use contrastive learning frameworks to model implicit representations of different degradation types, e.g., DASR [11], IDMBSR [15]. They assume that the degradations contained in different patches within the same image are consistent, while the degradation is different between patches in different images. Then, the implicit form of degradation representation is learned by pulling the positive samples closer together and pushing the negative samples farther apart. These methods initially explore the computational flow of implicit degradation estimation based on contrastive learning. However, due to the lack of systematic constraints, the implicit degradation representation learning is difficult to converge along the task-related direction and is easy to be interfered by content features such as background and appearance, resulting in the lack of sufficient discrimination of the extracted degradation features, which in turn affects the performance of image super-resolution.

This work was supported by the Beijing Natural Science Foundation(No.4234086); the Natural Science Foundation of China(No. 62192782). The work was completed during the internship in the Institute of Automation, Chinese Academy of Sciences (CASIA). (Corresponding author: Bo Wang)

Jiang Yuan and Ji Ma are with the State Key Laboratory of Multimodal Artificial Intelligence Systems, CASIA, Beijing 100190, China; School of Control and Computer Engineering, North China Electric Power University, Beijing 102206, China. E-mail: yuanj@ncepu.edu.cn; maji@ncepu.edu.cn. (The two authors contribute equally to this work)

Bo Wang is with the State Key Laboratory of Multimodal Artificial Intelligence Systems, CASIA, Beijing 100190, China. E-mail: wangbo@ia.ac.cn.

Weiming Hu is with State Key Laboratory of Multimodal Artificial Intelligence Systems, CASIA, Beijing 100190, China; School of Artificial Intelligence, University of Chinese Academy of Sciences, Beijing 100049, China; School of Information Science and Technology, ShanghaiTech University, Shanghai, 201210, China. Email: wmhu@nlpr.ia.ac.cn.

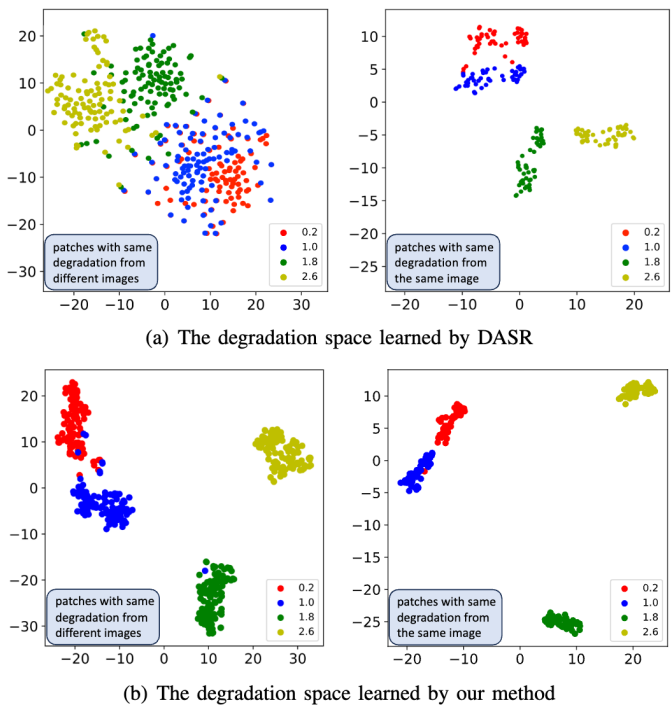

Fig. 1. The t-SNE [17] plots of degradation representation distributions with 4 different kernel widths on the DIV2K [16] val-set.

To explain the problem intuitively, we reproduce the classical contrastive learning-based DASR [11] on DIV2K val-set [16]. Images are first degraded using 4 different blur kernels (with kernel widths of 0.2, 1.0, 1.8, 2.6). Then patches are sampled from degraded images and their degradation representations are extracted by DASR. Finally, we visualize the distribution of degradation representations by t-SNE [17]. As is shown in Fig. 1(a), for degradation space learned by DASR, if the patches with the same degradation come from different images (Fig. 1(a) left), the sample distribution of each degradation is dispersed and lack discriminability between different degradations. But if the patches with the same degradation come from the same image (Fig. 1(a) right), the aggregation degree of representations for different degradations is significantly improved. This shows that the degradation space is seriously affected by content-related features.

To comprehensively improve the quality of implicit degradation modeling and SR performance under contrastive learning-based paradigm, we propose a new Content-decoupled Contrastive Learning-based image blind super-resolution (CdCL) framework (Fig. 3), which focuses on improving the implicit degradation estimator and degradation adaptation module. In the stage of degradation modeling, by optimizing the sampling strategy, we introduce the negative-free contrastive learning [18], [19] for the first time to train the implicit degradation estimator, called Content-decoupled Implicit Degradation Modeling (CdIDM). As shown in Fig. 2, CdIDM aims to decouple content and degradation features, thus improving the purity and discriminability of the learned degradation representations. Specifically, instead of constructing positive pairs based on different transformations of the same image, we design a cyclic shift sampling strategy, in which patches with the same degradation from different images are sampled to construct positive pairs, thus avoiding the positive-pairing between patches with similar content, and data augmentation is achieved by continuously pairing these patches in a cyclic manner. Then, we use divide-combine operation during embedding to increase the number of positive samples of the same degradation, thus further expanding the data diversity of each degradation category in the training stage. As shown in Fig. 1(b), regardless of the settings, the degradation space learned by our method achieves better intra-class high aggregation and inter-class high discrimination, proving that our method can learn a more discriminative feature space, where the sample clustering mainly relies on degradation information.

After obtaining high-quality degradation representations, how to effectively incorporate them into LR image features to guide the super-resolving process is also very critical, commonly known as degradation adaptation. To this end, we design a more lightweight Detail-aware implicit degradation adapting module (DaIDAM), which aims to enhance the adaptation network's perception of image details by incorporating LR features into the modulation process, thereby more effectively and efficiently adapt the implicit degradation representation into LR image features to guide the SR process. As shown in Fig. 3 and 4, DaIDAM consists of many detail-aware degradation adaptation unit (DaDAU) as basic components in series. Specifically, the degradation adaptation group (DAG) is first constructed by stacking a series of DaDAUs and convolution blocks, and then the DaIDAM is formed by stacking a series of DAGs and convolution blocks. Inside DaDAU, for spatial-level adaptation, instead of modulating degradation representations through continuous parameter-heavy FC layers [11], we use two simple convolution layers, significantly reducing the number of parameters and computation. At the channel-level, we incorporate LR features into the channel-wise degradation representation modulation to enhance the adaptability of DaDAU to specific image characteristics, thereby improving the fusion effect of degradation representations and LR features. Subsequent experiments validate the advantages of DaDAU. Without stacking a very deep adaptation network, our approach also achieves highly competitive SR results.

This paper constructs a new contrastive learning-based blind SR framework, whose main contributions include:

1) We propose a Content-decoupled Implicit Degradation Modeling (CdIDM) technique, in which a new cyclic shift sampling strategy is designed to successfully introduce negative-free contrastive learning into the training of the implicit degradation estimator. CdIDM avoids the interference of content-related features during implicit degradation modeling from the data perspective, extracting more discriminative degradation representations.

2) We propose a Detail-aware Implicit Degradation Adaptation Module (DaIDAM) that improves the fusion effect between degradation representations and LR features by enhancing the adaptation network's perception of image details rather than stacking adaptation blocks, significantly reducing the overall model complexity.

3) We conduct extensive experiments on popular benchmarks. The results show that our method achieves highly competitive quantitative and qualitative results with very low model complexity under various degradation settings, validating the importance of learning discriminative degradation representations and efficient degradation adaptation for developing lightweight blind SR tools.

## II. RELATED WORK

*A. Non-Blind Image Super-Resolution*

As a pioneering work, SRCNN [1] was the first CNN-based SR method that outperforms previous conventional SR methods with higher efficiency and better performance using only three convolutional layers. Since then, deep learning-based SR methods [4], [5], [20], [21] have been widely studied and become the dominant SR method. VDSR [22] introduced residual modules [23] on the basis of SRCNN, and proposed a very deep SR network, which significantly improves the SR effect. To further improve the performance, RCAN [2] incorporated a channel attention mechanism [24] that adjust the importance of different feature channels in the SR network. SwinIR [25] introduced the Transformer [26] into the SR network, which improved the SR effect by establishing a larger receptive field dependence than the CNN architecture. However, since the above SR methods assume that the degradation operation is a bicubic downsampling, they usually exhibit poor performance when the test degradation is inconsistent with the expected hypothesis. To this end, some methods attach a specific degradation parameter when inputting LR images into the SR network to guide the super-resolution process. For example, SRMD [8] combined the reshaped blur kernel with the LR image as input and can generated different SR results based on the provided blur kernel. UDVD [27] also used blur kernels as an additional input to SR process and employs pixel-by-pixel dynamic convolution to more efficiently deal with changing degradation within the image. However, degradation in the real-world is often unknown, so the application scenarios of these SR methods are still limited.

*B. Blind Image Super-Resolution*

To achieve high-quality SR in the case of unknown and mixed degradation, researchers have begun to study blind SR techniques [9], [10], [13], which uses a learnable degradation estimator to extract degradation information from LR images instead of manually setting the degradation parameter. The accuracy of the estimated degradation representation largely determines the final SR performance. Early blind SR methods adopt an explicit degradation estimation strategy, which trains a degradation estimator to extract precise degradation parameters. IKC [9] adopted an iterative estimation method, continuously using the generated SR image to correct the degradation representation and applying it to the next iteration until satisfactory SR results are obtained. DAN [10] designed a two-branch network to predict blur kernel and SR image in parallel and alternately updated the two branches during training, thus achieving synchronous optimization of blur kernel estimation and SR image reconstruction. DCLS [28] designed a deep constrained least square filtering module that adaptively extracted deblurring features from LR images and fused them with LR features through a dual-path network. Despite some progress, the above explicit degradation estimation-based SR methods often require a large number of iterations to accurately calculate degradation parameters, resulting in time-consuming and difficult to handle various complex degradation in practice.

To improve the efficiency and effectiveness of blind SR, some methods try to implicitly represent the degradation information in the latent space without requiring specific degradation information as supervision. For example, DASR [11] learned a latent representation of degradation information through the unsupervised contrastive learning technique [29], which can handle arbitrary degradation scenario with better generalization. IDMBSR [15] enhanced DASR by leveraging the blur kernel width and the noise level as weakly supervised information to guide the training of the implicit degradation estimator. DAA [30] adopted an implicit degradation modeling technique based on ranking loss [31], in which the implicit degradation estimator is trained by optimizing LR image sequences with different degradation complexities. MRDA [12] employed the meta-learning technique [32] and a multi-stage training strategy to implicitly learn the degradation representation. KDSR [33] proposed an implicit degradation estimation framework based on knowledge distillation [34], in which HR images are used to assist in training the teacher network to model different degradation types, and the learned knowledge is passed on to the student network for implicit degradation estimation in the inference stage. DSAT [35] was the first to integrate a contrastive learning-based implicit degradation learning framework with a Transformer-based SR network. CDFormer [36] designed the first diffusion-based implicit degradation representation estimator and constructed a very large SR network by stacking Transformer blocks [37].

## III. METHOD

*A. Overview*

Given an original low-resolution (LR) image, our goal is to restore the corresponding high-resolution (HR) version in the presence of unknown degradation information. The overall architecture is shown in Fig. 3, which mainly consists of four parts: **Image feature extraction module**, which adopts a simple 3×3 convolution layer, is used to extract low-level details of the image (such as texture, color, etc.); **Implicit degradation estimation module** consists of a degradation estimator and a degradation regulator. The former is a 6-layer convolutional network trained using the proposed CdIDM technique to extract the implicit degradation representation of any LR image. The latter includes a channel transformation branch and a spatial transformation branch, which are used to further generate channel-wise degradation representation and spatial-wise degradation representation respectively based on the output of the estimator; **Implicit degradation adaptation module** is constructed according to the proposed detail-aware implicit degradation adapting scheme, which is used to incorporate implicit degradation representations into specific LR image features from both channel and spatial perspectives through a series of DaDAUs and DAGs. **Image reconstruction module** is used to generate a content-consistent HR image based on LR image features fused with degradation information. Here we use Upscaler [7] directly. The training of our method is divided into two stages. The first stage is to pretrain

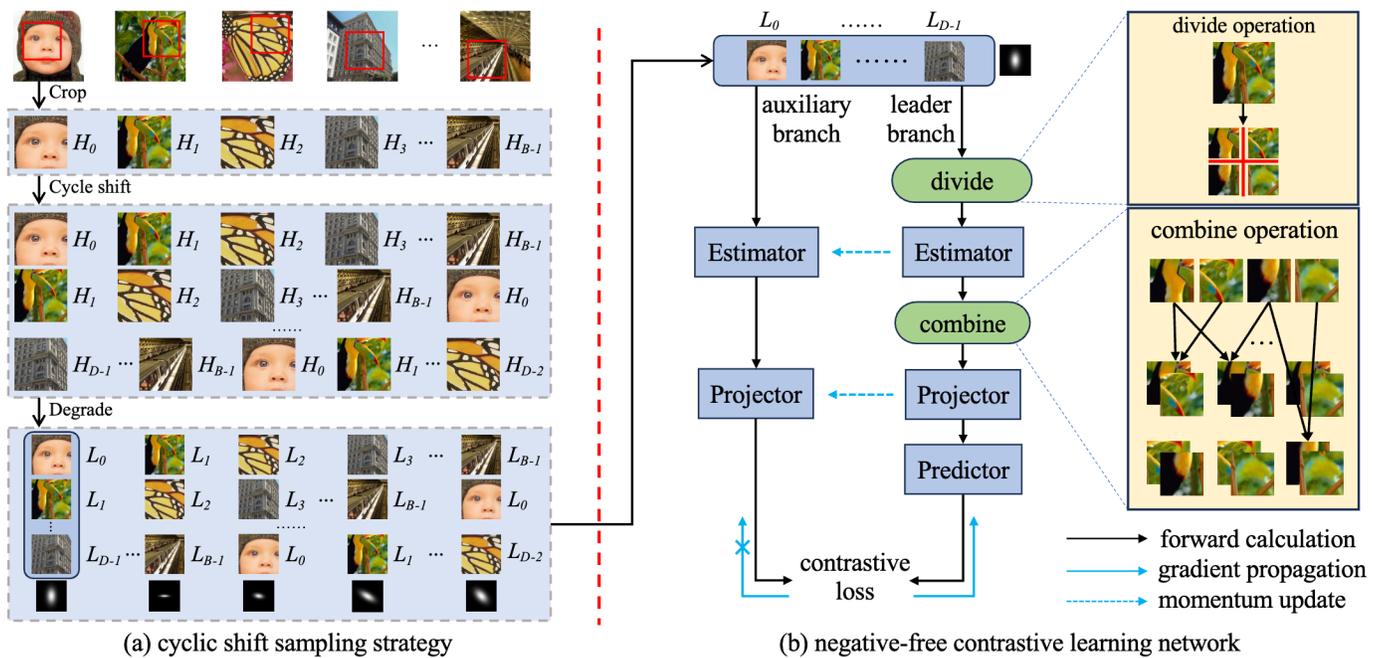

Fig. 2. The illustration of the proposed content-decouple implicit degradation modeling.

the implicit degradation estimator, and the second stage is to train the whole SR network.

### B. Content-decoupled Implicit Degradation Modeling

To learn more pure and discriminative degradation representations, we propose a content-decouple implicit degradation modeling framework (as shown in Fig. 2), which consists of a novel cyclic shift sampling strategy and an end-to-end two-branch negative-free contrastive learning network.

*1) Cyclic shift sampling*

A typical negative-free contrastive learning framework uses different transformations of the same sample to form positive pairs, and relies on content similarity to learn a feature space, which is exactly what implicit degradation modeling must avoid. To this end, we design a new cyclic shift sampling strategy to improve the above framework and ensure the decoupling of the learned degradation representations from the content features at the data level. Specifically, as shown in Fig. 2(a), given a set of HR images $\mathcal{I} = \{I_0, \ldots, I_{B-1}\}$, first, we randomly sample a patch from each HR image to form a HR patch sequence $\mathcal{S}_0 = \{S_0, \ldots, S_{B-1}\}$. Second, we cyclically shift the sequence with a step size of 1 to obtain a new sequence $\mathcal{S}_1 = \{S_1, \ldots, S_{B-1}, S_0\}$. This operation is performed continuously $D-1$ times, resulting in $D-1$ HR patch sequences. Third, we combine the original sequence $\mathcal{S}_0$ with all the sequences obtained by cyclic shifting to form a HR patch matrix $\mathcal{M} = \{\mathcal{S}_0; \ldots; \mathcal{S}_{D-1}\}$ with $D$ rows and $B$ columns. Finally, we apply a different degradation parameters to each column in $\mathcal{M}$ to obtain the corresponding LR version of the HR patch. That is, the degradation of patches in each column is consistent, and the degradation of patches in different columns is different. In this way, we obtain $B$ positive sample sets (each with the same degradation), which can be used to make positive pairs for subsequent contrastive learning. Meanwhile, we also get $D \times B$ HR/LR patch pairs, which can be used for the training of the SR network. CdIDM ensures that all samples within the same set are patches from different LR images subjected to the same degradation. This data-level arrangement minimizes the likelihood of high similarity in image content among positive samples.

*2) Implicit degradation modeling*

The framework for modeling degradation representations is shown in Fig. 2(b), consists of two branches: a leader branch and an auxiliary branch. The leader branch consists of an estimator, a projector, and a predictor. The estimator is a six-layer convolutional network. The projector and predictor are two-layer fully connected networks with batch-normalization [38] and GELU activation [39]. The auxiliary branch consists only of an estimator and a projector, and their structure are the same as that in the leader branch.

During training, for the leader branch, given $B$ positive sample sets (each consisting of $D$ LR patches with the same degradation), each LR patch is first uniformly divided into 4 non-overlapping local blocks and sent to the estimator for encoding. Then, the encoded features are combined into 6 different pairs, and the features of each pair are averaged separately. Finally, all features are sequentially fed into the projector and predictor for continuous feature transformation. This divide-combine technique can further expand the sample diversity for each degradation category. The output of the leader branch is defined as $\mathcal{O} \in \mathbb{R}^{B \times D \times 6 \times 256}$, where 256 is the dimension of the implicit degradation representation. For the auxiliary branch, the LR patch is directly sent to the estimator for encoding. Then the encoded features are fed into the projector for transformation. The output of the auxiliary branch is defined as $\mathcal{T} \in \mathbb{R}^{B \times D \times 256}$.

The parameter update of the contrastive learning network is divided into two steps. The first step is to update the parameters of the leader branch by utilizing the contrastive loss between the outputs of the leader branch and the auxiliary

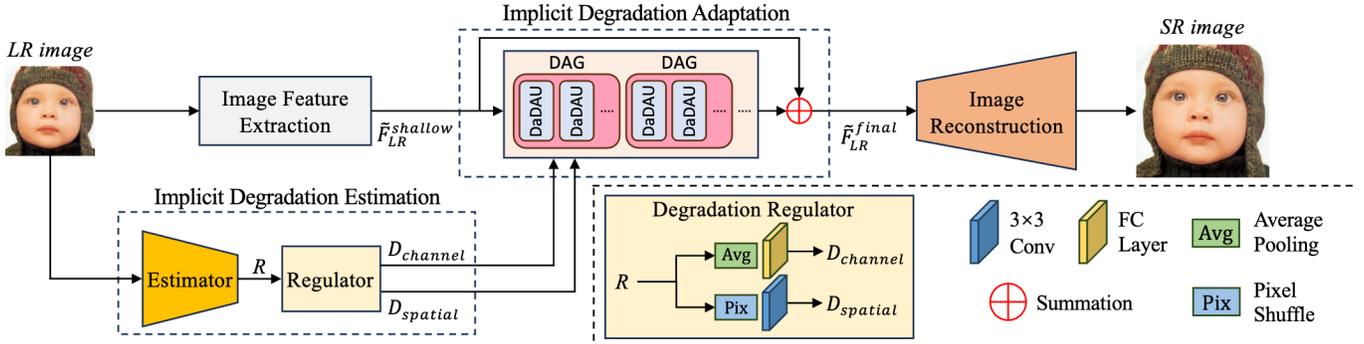

Fig. 3. The architecture of our proposed Content-decoupled Contrastive Learning-based Blind Image SR Network (CdCL).

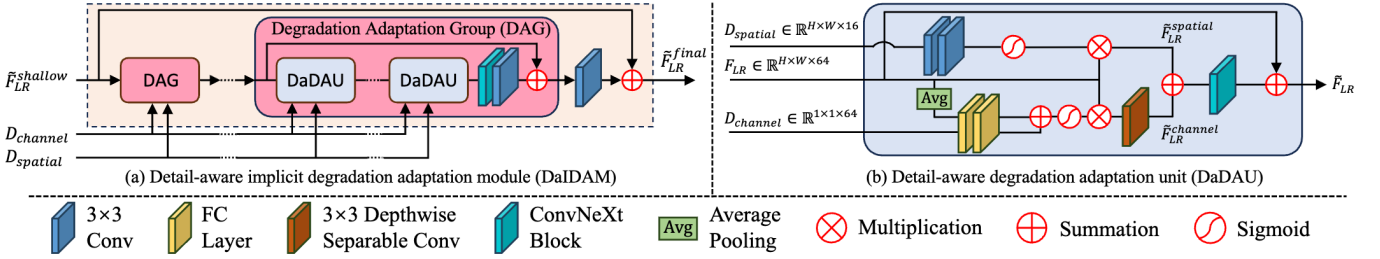

Fig. 4. The detailed structure of the DaDAU, DAG and DaIDAM.

branch. The loss function is defined as follows:

$$L_{contr} = \frac{1}{6D(D-1)} \sum_{i=1}^{D} \sum_{k=1}^{6} \sum_{j=1}^{D} Sim(\mathcal{O}^{i,k}, \mathcal{T}^j), (i \neq j), \quad (1)$$

where the distance function $Sim(\cdot)$ uses InfoNCE [40]:

$$Sim(\mathcal{O}^{i,k}, \mathcal{T}^j) = \frac{1}{B} \sum_{n=1}^{B} -\log \frac{\exp(\mathcal{O}_n^{i,k} \cdot \mathcal{T}_n^j / \tau)}{\sum_{m=1}^{B} \exp(\mathcal{O}_n^{i,k} \cdot \mathcal{T}_m^j / \tau)}. \quad (2)$$

Based on the constructed positive sample sets, contrastive loss drives the estimator to pull together content-different but degradation-similar patches and push apart content-similar but degradation-different patches, so as to focus on capturing degradation-related discriminative features.

The second step is to use the momentum update technique to update the parameters of the auxiliary branch, as follows:

$$\begin{aligned} \theta_{\mathcal{T}}^{Est} &\leftarrow (1-\alpha)\theta_{\mathcal{T}}^{Est} + \alpha\theta_{\mathcal{O}}^{Est} \\ \theta_{\mathcal{T}}^{Pro} &\leftarrow (1-\alpha)\theta_{\mathcal{T}}^{Pro} + \alpha\theta_{\mathcal{O}}^{Pro}, \end{aligned} \quad (3)$$

where $\theta_{\mathcal{O}}^{Est}$ and $\theta_{\mathcal{O}}^{Pro}$ are the parameters of the estimator and projector in the leader branch respectively, $\theta_{\mathcal{T}}^{Est}$ and $\theta_{\mathcal{T}}^{Pro}$ are the parameters of the estimator and projector in the auxiliary branch respectively, $\alpha \in [0, 1]$ is the momentum coefficient. This asymmetric design helps avoid training crashes [18].

*3) Degradation representation regulating*

After training, we use the estimator in the leader branch to extract the degradation representation of the LR image, and use the regulator to extend it from both channel and spatial perspectives. Specially, given an LR image $I_{LR} \in \mathbb{R}^{H \times W \times 3}$ and the extracted degradation representation $R \in \mathbb{R}^{\frac{H}{4} \times \frac{W}{4} \times 256}$, the channel transformation branch performs global average pooling on $R$ and sends the pooled result to a FC layer for non-linear transformation, obtaining the channel-wise degradation representation $D_{channel} \in \mathbb{R}^{1 \times 1 \times 64}$. The spatial transformation branch uses the pixel-shuffle layer [7] to upsample $R$ to the same size as $I_{LR}$ and sends the upsampled featuremaps into two 3×3 convolution layers to obtain the spatial-wise degradation representation $D_{spatial} \in \mathbb{R}^{H \times W \times 16}$.

*C. Detail-aware Implicit Degradation Adapting*

After obtaining degradation representations and LR image features, the adaptation between them is the next key step, which determines the HR reconstruction effect. To this end, we propose a detail-aware degradation adaptation unit (DaDAU) as the basic unit, and gradually construct the degradation adaptation group (DAG) and the final DaIDAM, as shown in Fig. 4. Our method aims to improve the fusion effect of degradation representations and image features by enhancing the adaptation network's perception of image details rather than stacking adaptation blocks.

*1) Unit level adaptation*

DaDAU is used for unit-level adaptation, which uses a two-branch structure to modulate degradation information from both channel and spatial perspectives to fit the specific image details. As shown in Fig. 4(b). In the spatial adaptation branch, two 3×3 convolution layers are first used to modulate the spatial degradation representation $D_{spatial}$, followed by a Sigmoid function for normalization. Then the modulated degradation representation is fused into the current image features $F_{LR}$. This process can be expressed as:

$$\widetilde{F}_{LR}^{spatial} = \sigma(W_{3\times3}^{conv}(\tau(W_{3\times3}^{conv}(D_{spatial})))) \odot F_{LR}, \quad (4)$$

where $W_{3\times3}^{conv}$ is a 3×3 convolution, $\tau(\cdot)$ is the GELU activation function, $\sigma(\cdot)$ is the Sigmoid function, $\widetilde{F}_{LR}^{spatial}$ are image features fused with spatial-wise degradation representation. This design modulates the spatial-wise degradation representation through parameter-efficient convolutional layers, significantly reducing the number of parameters.

In the channel adaptation branch, to introduce $F_{LR}$ in the modulation of $D_{channel}$, the global pooling layer is firstly used to adjust the shape of $F_{LR}$. Secondly, $D_{channel}$ and $\text{GAP}(F_{LR})$ are fed into the same two-layer FC network, and their transformed results are summed. Then a Sigmoid function is used to normalize the modulated degradation representation. Finally, the normalized result is fused into $F_{LR}$, followed by a 3×3 depthwise separable convolution layer for nonlinear transformation. This process can be expressed as:

$$\widetilde{F}_{LR}^{channel} = W_{3\times3}^{dwconv}((\sigma(W^{FC}(\tau(W^{FC}(D_{channel})))) + W^{FC}(\tau(W^{FC}(GAP(F_{LR}))))) \odot F_{LR}), \quad (5)$$

where $W_{3\times3}^{dwconv}$ is a 3×3 depthwise separable convolution layer, $W^{FC}$ is a FC layer, $GAP(\cdot)$ is a global average pooling operation, $\widetilde{F}_{LR}^{channel}$ are image features fused with channel-wise degradation representation. This design incorporates current LR features to modulate the degradation representation, thereby better adapting to input conditions.

After the calculation of the two branches, $\widetilde{F}_{LR}^{spatial}$ and $\widetilde{F}_{LR}^{channel}$ are element-wise summed and fed into a ConvNeXt block [41] for further fusion. Finally, $F_{LR}$ is fused using a residual link to recover the high-frequency detail information. This process can be expressed as:

$$\widetilde{F}_{LR} = W^{CXt}(\widetilde{F}_{LR}^{spatial} + \widetilde{F}_{LR}^{channel}) + F_{LR}, \quad (6)$$

where $W^{CXt}$ is a ConvNeXt block consisting of a 7×7 depthwise separable convolution layer and two 1×1 convolution layers. $\widetilde{F}_{LR}$ is the degradation-adapted image features.

*2) Group level adaptation*

Based on DaDAU, we construct a degradation adaptation group (DAG), which consists of a series of DaDAUs, a ConvNeXt block and a 3×3 convolution layer. As shown in the red rounded rectangle in Fig. 4(a). LR image features and spatial/channel degradation representations are sequentially fed into each component for processing, and a residual link is added between the input and output of a DAG.

*3) Module level adaptation*

Based on DAG, we built the detail-aware implicit degradation adaptation module (DaIDAM), as shown in Fig. 4(a), which consists of a series of DAGs and a 3×3 convolution layer in series. Similarly, we add a residual link between the input $\widetilde{F}_{LR}^{shallow} \in \mathbb{R}^{H\times W\times 64}$ and output of the DaIDAM to restore image detail features. After adaptation processing, we obtain the LR image features fully fused with degradation information, defined as $\widetilde{F}_{LR}^{final} \in \mathbb{R}^{H\times W\times 64}$.

### D. Implicit Degradation-guided Blind Super-resolving

*1) Image reconstruction*

After completing the adaptation of implicit degradation representation and LR image features, our CdCL network use Upscaler [7] to reconstruct the corresponding HR version $\hat{I}_{HR}$ of the input LR image. Upscaler is a learnable upsampling module composed of several convolutional layers and pixel shuffle layers, and its specific structure is determined by the scaling factor of the SR task. Given the LR image features $\widetilde{F}_{LR}^{final}$, the reconstruction process can be expressed as:

$$\hat{I}_{HR} = Upscaler(\widetilde{F}_{LR}^{final}). \quad (7)$$

*2) Model training*

During training, we use the L1 loss to update the parameters of all modules of the proposed blind SR network:

$$L_{L1} = \left\| \hat{I}_{HR} - I_{HR} \right\|_1, \quad (8)$$

where $I_{HR}$ is the HR version of the input LR image in GT.

## IV. EXPERIMENTS

### A. Experimental Setup

*1) Datasets preparation*

Following [9], [11], [15], we combine 800 images from DIV2K [16] and 2650 images from Flickr2K [42] as the training dataset, and use 4 standard benchmarks (Set5 [43], Set14 [44], B100 [45] and Urban100 [46]) for quantitative and qualitative performance evaluation. Based on the above datasets, we use 3 different degradation settings to generate the LR/HR image pairs needed for training and testing. For degradation processing, we follow the classical image degradation model [47], which can be expressed as follows:

$$I_{LR} = [(I_{HR} \otimes k) \downarrow_s + n]_{JPEG}, \quad (9)$$

where $I_{HR}$ is the original HR image, $I_{LR}$ is the degraded LR image, $\otimes$ denotes the convolution operation, $k$ denotes the blur kernel, $\downarrow_s$ denotes the downsampling with the scaling factor $s$, $n$ denotes the Gaussian white noise, and $JPEG$ denotes the JPEG compression operation. In addition, we qualitatively evaluate the generalization of the proposed method in the real-world dataset RealWorld38 [48].

*Degradation setting 1* contains only isotropic Gaussian blur kernels without noise and JPEG compression operation. The Gaussian kernel size is fixed to 21×21, the kernel width range is set to [0.2, 4.0]. The scaling factor is 3 or 4.

*Degradation setting 2* considers anisotropic Gaussian blur kernels and Gaussian white noise without JPEG compression. The anisotropic Gaussian blur kernel characterized by a Gaussian probability density function $N(0, \Sigma)$ and the covariance matrix $\Sigma$ is determined by two random eigenvalues $\lambda_1, \lambda_2 \sim U(0.2, 4)$ and a random rotation angle $\theta \sim U(0, \pi)$. The Gaussian kernel size is fixed to 21×21, the Gaussian white noise range is set to [0, 25]. The scaling factor is 4.

*Degradation setting 3* covers all degradation categories, including isotropic Gaussian blur kernels, Gaussian white noise and JPEG compression. All three degradations occur with a probability of 50%. The Gaussian kernel size remains fixed at 21×21, the kernel width ranges from [0.1, 3.0], the noise ranges from [1, 30], the JPEG compression quality factor ranges from [40, 95]. The scaling factor is 4.

*2) Metrics for evaluation*

We use 3 metrics to assess the SR quality. Peak signal-to-noise ratio (PSNR) calculates the pixel-level similarity between SR and GT images. Structural similarity index (SSIM) [49] evaluates the low-level similarity between SR and GT images by comparing the luminance, contrast, and structure. Learned perceptual image patch similarity (LPIPS) [50] utilizes a pretrained CNN to calculates the content-level similarity. We use FLOPs as the unit of computational complexity, and all models are tested based on 180×320 LR size.

TABLE I
×4 SR QUANTITATIVE COMPARISON UNDER DEGRADATION SETTING 1.

| Method | Param(M) | FLOPs(G) | Dataset | Set5 | | | Set14 | | | B100 | | | Urban100 | | |
|---|---|---|---|---|---|---|---|---|---|---|---|---|---|---|---|
| | | | Kernel Width | 1.2 | 2.4 | 3.6 | 1.2 | 2.4 | 3.6 | 1.2 | 2.4 | 3.6 | 1.2 | 2.4 | 3.6 |
| Bicubic | — | — | PSNR | 27.69 | 25.99 | 24.45 | 25.60 | 24.39 | 23.25 | 25.58 | 24.67 | 23.80 | 22.73 | 21.76 | 20.84 |
| | | | SSIM | 0.7904 | 0.7351 | 0.6774 | 0.6820 | 0.6292 | 0.5811 | 0.6461 | 0.5982 | 0.5570 | 0.6341 | 0.5791 | 0.5305 |
| *Comparison of Explicit Degradation Blind SR Methods* | | | | | | | | | | | | | | | |
| IKC | 5.3 | 2528.03 | PSNR | 31.76 | 30.35 | 30.26 | 28.44 | 28.17 | 26.63 | 27.43 | 27.28 | 26.41 | 25.63 | 25.02 | 24.07 |
| | | | SSIM | 0.8870 | 0.8574 | 0.8584 | 0.7714 | 0.7606 | 0.7100 | 0.7240 | 0.7164 | 0.6854 | 0.7676 | 0.7427 | 0.7024 |
| DAN | 4.3 | 1098.33 | PSNR | 32.22 | 31.96 | 30.94 | 28.65 | 28.54 | 27.68 | 27.65 | 27.58 | 26.95 | 26.20 | 25.96 | 25.08 |
| | | | SSIM | 0.8959 | 0.8881 | 0.8659 | 0.7795 | 0.7724 | 0.7372 | 0.7343 | 0.7282 | 0.6950 | 0.7866 | 0.7758 | 0.7394 |
| DCLS | 13.6 | 436.57 | PSNR | 32.33 | 32.20 | 31.15 | 28.63 | 28.59 | 27.87 | 27.72 | 27.62 | 27.06 | 26.51 | 26.23 | 25.34 |
| | | | SSIM | 0.8956 | 0.8910 | 0.8689 | 0.7826 | 0.7761 | 0.7429 | 0.7380 | 0.7314 | 0.6998 | 0.7958 | 0.7841 | 0.7488 |
| CdCL-S | 2.9 | 91.68 | PSNR | 32.48 | 32.23 | 31.04 | 28.79 | 28.64 | 27.73 | 27.73 | 27.63 | 27.00 | 26.36 | 26.06 | 25.20 |
| | | | SSIM | 0.8972 | 0.8917 | 0.8682 | 0.7835 | 0.7763 | 0.7408 | 0.7381 | 0.7316 | 0.6984 | 0.7943 | 0.7824 | 0.7460 |
| *Comparison of CNN-based Implicit Degradation Blind SR Methods* | | | | | | | | | | | | | | | |
| DASR | 5.8 | 185.66 | PSNR | 31.92 | 31.75 | 30.59 | 28.44 | 28.28 | 27.45 | 27.52 | 27.43 | 26.83 | 25.69 | 25.44 | 24.66 |
| | | | SSIM | 0.8904 | 0.8855 | 0.8601 | 0.7731 | 0.7644 | 0.7293 | 0.7303 | 0.7203 | 0.6909 | 0.7700 | 0.7581 | 0.7222 |
| IDMBSR | 4.2 | — | PSNR | 31.90 | 31.78 | 30.68 | 28.50 | 28.36 | 27.60 | 27.58 | 27.51 | 26.90 | 25.91 | 25.68 | 24.91 |
| | | | SSIM | 0.8903 | 0.8860 | 0.8624 | 0.7770 | 0.7706 | 0.7364 | 0.7334 | 0.7278 | 0.6958 | 0.7780 | 0.7680 | 0.7330 |
| MRDA | 5.8 | 172.13 | PSNR | 32.36 | 32.11 | 30.89 | 28.67 | 28.57 | 27.62 | 27.67 | 27.58 | 26.91 | 26.26 | 26.02 | 25.08 |
| | | | SSIM | 0.8958 | 0.8900 | 0.8644 | 0.7804 | 0.7738 | 0.7344 | 0.7359 | 0.7301 | 0.6954 | 0.7889 | 0.7783 | 0.7380 |
| KDSR | 5.8 | 191.42 | PSNR | 32.34 | 32.13 | 31.02 | 28.66 | 28.55 | 27.81 | 27.67 | 27.59 | 26.97 | 26.29 | 26.05 | 25.20 |
| | | | SSIM | 0.8964 | 0.8915 | 0.8687 | 0.7865 | 0.7801 | 0.7467 | 0.7380 | 0.7320 | 0.6992 | 0.7899 | 0.7798 | 0.7455 |
| CdCL-S | 2.9 | 91.68 | PSNR | 32.48 | 32.23 | 31.04 | 28.79 | 28.64 | 27.73 | 27.73 | 27.63 | 27.00 | 26.36 | 26.06 | 25.20 |
| | | | SSIM | 0.8972 | 0.8917 | 0.8682 | 0.7835 | 0.7763 | 0.7408 | 0.7381 | 0.7316 | 0.6984 | 0.7943 | 0.7824 | 0.7460 |
| *Comparison of Transformer-based Implicit Degradation Blind SR Methods* | | | | | | | | | | | | | | | |
| DSAT | 15.6 | 803.94 | PSNR | 32.51 | 32.12 | 30.34 | 28.72 | 28.63 | 27.56 | 27.73 | 27.63 | 27.02 | 26.54 | 26.02 | 25.01 |
| | | | SSIM | 0.8982 | 0.8896 | 0.8535 | 0.7819 | 0.7728 | 0.7283 | 0.7396 | 0.7338 | 0.7008 | 0.7975 | 0.7788 | 0.7372 |
| CDFormer-S | 11.9 | 634.53 | PSNR | 32.36 | 32.14 | 31.07 | 28.77 | 28.60 | 27.67 | 27.74 | 27.64 | 27.00 | 26.39 | 26.09 | 25.19 |
| | | | SSIM | 0.8959 | 0.8900 | 0.8677 | 0.7829 | 0.7740 | 0.7369 | 0.7385 | 0.7307 | 0.6969 | 0.7921 | 0.7797 | 0.7425 |
| CdCL-L | 5.9 | 201.68 | PSNR | 32.52 | 32.25 | 31.12 | 28.84 | 28.71 | 27.85 | 27.79 | 27.70 | 27.05 | 26.58 | 26.28 | 25.37 |
| | | | SSIM | 0.8983 | 0.8930 | 0.8694 | 0.7843 | 0.7774 | 0.7431 | 0.7403 | 0.7343 | 0.7010 | 0.7983 | 0.7868 | 0.7502 |

*3) Details for model training*

For pre-processing, random flipping and random rotation are used to further augment original HR images. During entire training, the batchsize $B$ is set to 64, the patch size is set to 64×64, and the number of cycle shifts $D$ is set to 4. AdamW optimizer [51] is used for parameters updating. In the implicit degradation modeling stage, the degradation estimator is first pretrained using $L_{contr}$ for 100 epochs, with an initial learning rate set to 1e-3 and reduced to 2e-4 at the 60th epoch. In the implicit degradation-guided blind SR network learning stage, the degradation estimator and other modules of the SR network were jointly trained for 600 epochs by summing $L_{contr}$ and $L_{L1}$, with the learning rate gradually adjusted from 2e-4 to 1e-6 using the cosine annealing strategy [52]. Note that we train two versions for subsequent experiments. The small version CdCL-S includes an image feature extractor with 64 output channels and a DaIDAM with 6 DAGs. The large version CdCL-L includes an image feature extractor with 96 output channels and a DaIDAM with 8 DAGs.

*B. Comparison with Previous Methods*

We conduct comparative experiments under 3 degradation settings. Comparison SR methods include IKC [9], DAN [10], DCLS [28], DASR [11], IDMBSR [15], MRDA [12], KDSR [33], DSAT [35] and CDFormer [36]. Note that among all competitors, IKC [9], DAN [10] and DCLS [28] are explicit estimation based blind SR methods, while others are implicit estimation based blind SR methods. Comparative perspectives include the quality of degradation representation, model complexity, quantitative and qualitative SR effect. It should be noted that we chose the smaller version of CDFormer, i.e. CDFormer-S, for a relatively fair comparison.

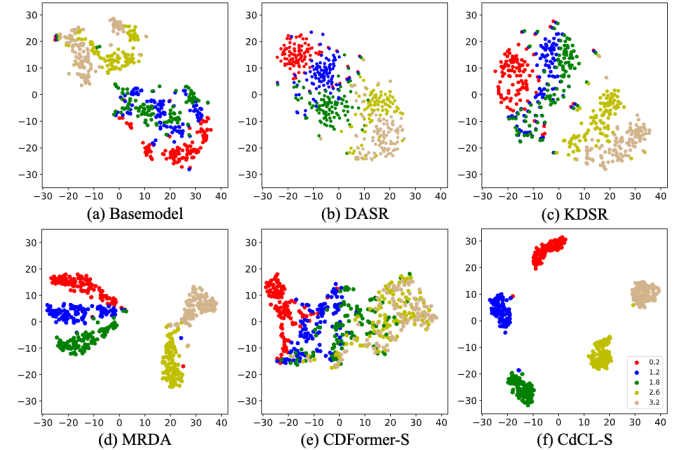

Fig. 5. The t-SNE [17] plots of degradation representation distributions with 5 different kernel widths on the DIV2K [16] val-set.

*1) Experiments in Degradation Setting 1*

**Quality of degradation representation.** We generate LR images from DIV2K val-set [16] under *Degradation Setting 1* with scaling factor 4 and feed them to Basemodel, DASR [11], MRDA [12], KDSR [33], CDFormer-S [36] and CdCL-S to generate corresponding degradation representations. The basemodel is an blind SR network without extra implicit degradation representation learning, MRDA is meta-learning based, KDSR is knowledge distillation based, DASR is contrastive learning based, and CDFormer-S is diffusion based. Then we use t-SNE technique to visualize the output implicit degradation representations. Fig. 5(a) shows that it is difficult for the estimator to construct a discriminative degradation space without clear degradation learning. As can be seen from Fig. 5(b)-(f), our method achieves better degradation-based clustering, where representations of the same degradation are closer to each other and representations of different degradations are more dispersed. The comparison with DASR proves that excluding task-irrelevant factors helps to learn more discriminative degradation representations. The comparison with MRDA, KDSR and CDFormer-S proves that contrastive learning-based paradigm is still a better solution in implicit degradation modeling, and its training is easier.

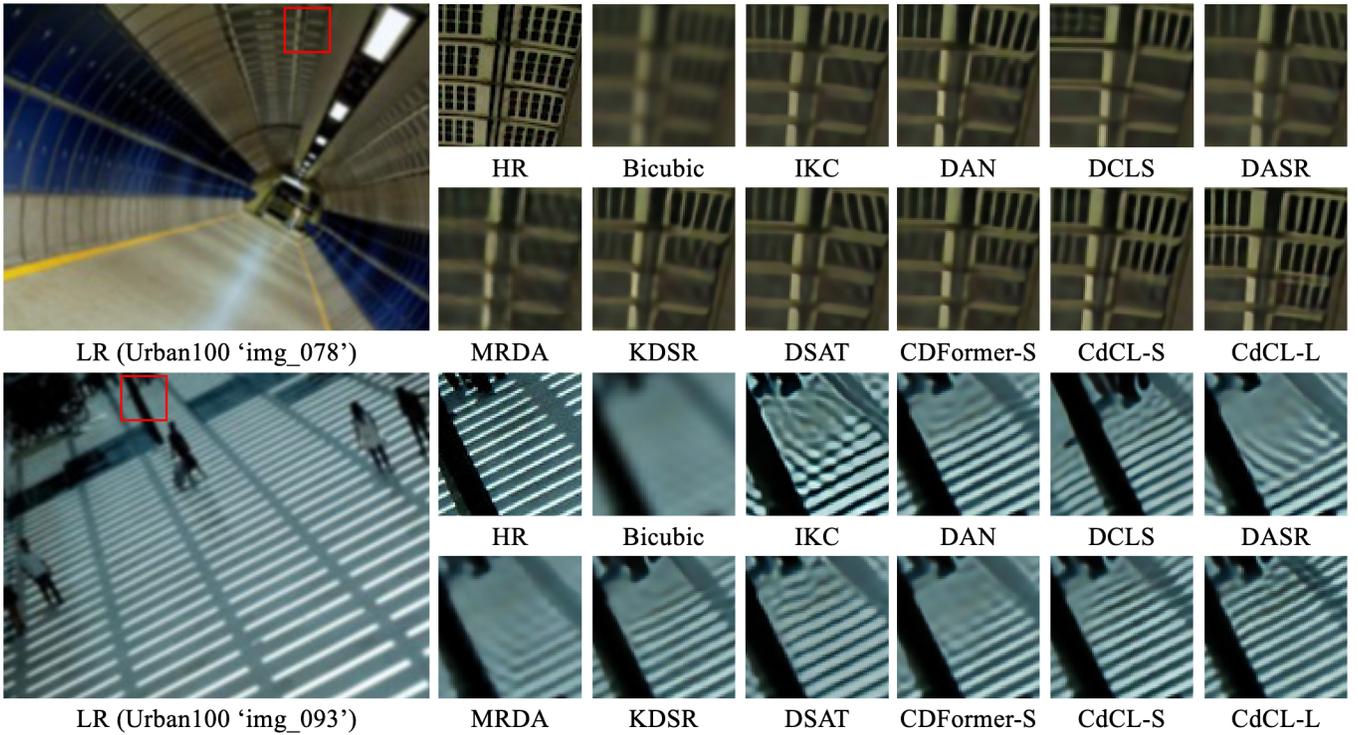

Fig. 6. The "$img\_078$" and "$img\_093$" on Urban100, blur kernel width is 2.4, ×4 SR result.

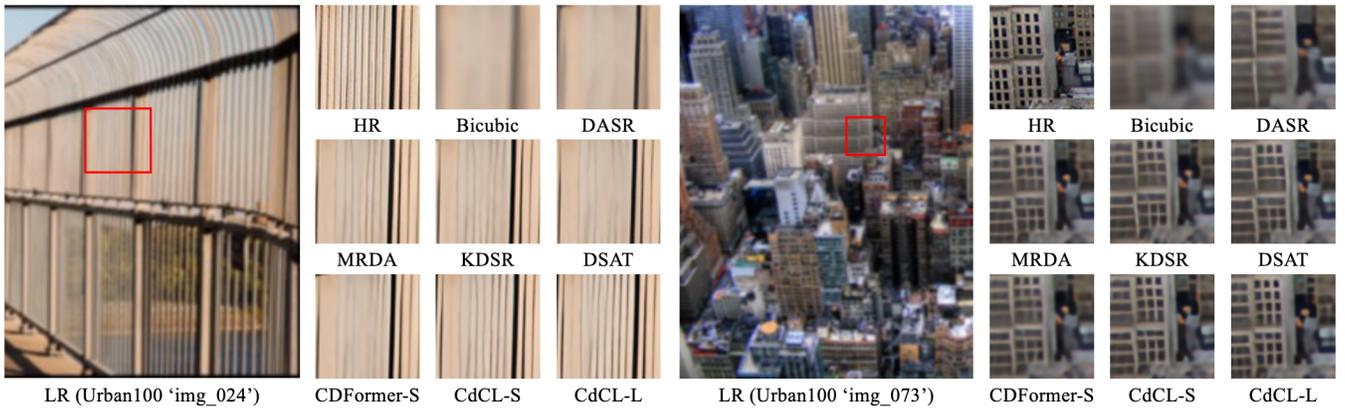

Fig. 7. ×3 SR results of "$img\_024$" and "$img\_073$" on Urban100 with blur kernel width 2.4.

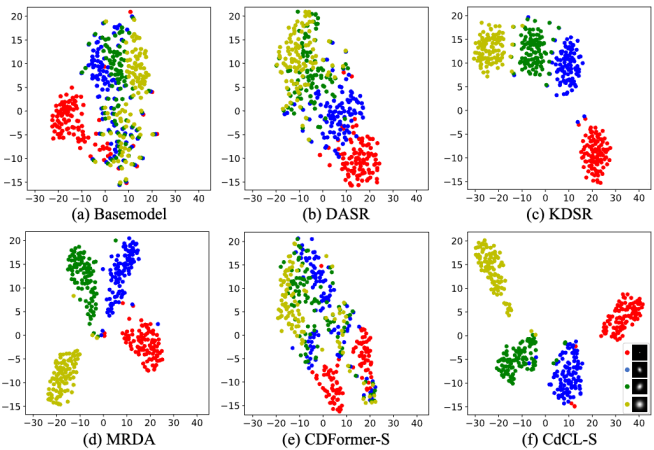

Fig. 8. The t-SNE [17] plots of degradation representation distributions with 4 different anisotropic kernels with noise level 4 on B100.

**Quantitative evaluation.** For scaling factor 4 with kernel widths {1.2, 2.4, 3.6}, the quantitative comparison results are shown in Table I. (1) Compared with the explicit estimation based blind SR methods, our approach achieves comparable or even superior results while significantly reducing computational cost and the number of parameters. For example, our method achieves results comparable to DCLS [28] with only 21% of its number of parameters and computational cost. With blur kernel widths of 1.2, 2.4, 3.6 and four benchmarks, our method outperforms DCLS in 6 out of 12 degradation

TABLE II
×3 SR QUANTITATIVE COMPARISON UNDER DEGRADATION SETTING 1. ↑ DENOTES THE HIGHER THE BETTER, ↓ DENOTES THE LOWER THE BETTER.

| Method | Param(M) | FLOPs(G) | Dataset | Set5 | | | Set14 | | | B100 | | | Urban100 | | |
|---|---|---|---|---|---|---|---|---|---|---|---|---|---|---|---|
| | | | Width | 0.8 | 1.6 | 2.4 | 0.8 | 1.6 | 2.4 | 0.8 | 1.6 | 2.4 | 0.8 | 1.6 | 2.4 |
| *Comparison of CNN-based Implicit Degradation Blind SR Methods* | | | | | | | | | | | | | | | |
| DASR | 5.7 | 160.37 | PSNR↑ SSIM↑ LPIPS↓ | 34.08 0.9231 0.0820 | 33.94 0.9200 0.0863 | 32.50 0.8983 0.1020 | 30.03 0.8309 0.1308 | 29.88 0.8228 0.1360 | 29.21 0.8003 0.1485 | 28.90 0.7940 0.1318 | 28.82 0.7893 0.1350 | 28.23 0.7664 0.1456 | 27.42 0.8324 0.1168 | 27.12 0.8235 0.1233 | 26.33 0.7952 0.1422 |
| MRDA | 5.6 | 145.69 | PSNR↑ SSIM↑ LPIPS↓ | 34.28 0.9258 0.0810 | 34.14 0.9225 0.0846 | 33.35 0.9087 0.0960 | 30.22 0.8371 0.1269 | 30.20 0.8331 0.1303 | 29.60 0.8104 0.1414 | 29.08 0.8015 0.1280 | 29.04 0.7984 0.1315 | 28.56 0.7752 0.1404 | 27.82 0.8431 0.1098 | 27.63 0.8376 0.1139 | 26.94 0.8157 0.1272 |
| KDSR | 5.6 | 202.83 | PSNR↑ SSIM↑ LPIPS↓ | 34.45 0.9261 0.0809 | 34.18 0.9225 0.0841 | 33.27 0.9083 0.0960 | 30.30 0.8382 0.1255 | 30.25 0.8344 0.1289 | 29.66 0.8134 0.1407 | 29.13 0.8028 0.1270 | 29.10 0.7994 0.1301 | 28.62 0.7763 0.1405 | 28.04 0.8484 0.1059 | 27.83 0.8422 0.1104 | 27.10 0.8191 0.1248 |
| CdCL-S | 2.9 | 67.57 | PSNR↑ SSIM↑ LPIPS↓ | 34.45 0.9265 0.0803 | 34.20 0.9228 0.0837 | 33.36 0.9088 0.0958 | 30.45 0.8405 0.1247 | 30.42 0.8370 0.1274 | 29.78 0.8146 0.1399 | 29.17 0.8036 0.1270 | 29.16 0.8009 0.1300 | 28.65 0.7771 0.1403 | 28.22 0.8526 0.1039 | 28.01 0.8462 0.1085 | 27.26 0.8232 0.1233 |
| *Comparison of Transformer-based Implicit Degradation Blind SR Methods* | | | | | | | | | | | | | | | |
| DSAT | 15.0 | 779.35 | PSNR↑ SSIM↑ LPIPS↓ | 34.56 0.9275 0.0779 | 33.96 0.9204 0.0843 | 32.01 0.8865 0.1096 | 30.49 0.8413 0.1230 | 30.25 0.8339 0.1296 | 29.10 0.7989 0.1473 | 29.28 0.8056 0.1249 | 29.09 0.7984 0.1315 | 28.37 0.7665 0.1422 | 28.35 0.8549 0.1028 | 27.92 0.8436 0.1097 | 26.68 0.8088 0.1320 |
| CDFormer-S | 11.1 | 610.43 | PSNR↑ SSIM↑ LPIPS↓ | 34.20 0.9261 0.0820 | 33.95 0.9224 0.0862 | 33.13 0.9083 0.0985 | 30.26 0.8386 0.1273 | 30.17 0.8334 0.1314 | 29.54 0.8099 0.1433 | 29.06 0.8010 0.1291 | 29.01 0.7974 0.1325 | 28.52 0.7742 0.1419 | 27.80 0.8439 0.1109 | 27.59 0.8373 0.1160 | 26.90 0.8137 0.1305 |
| CdCL-L | 5.9 | 148.01 | PSNR↑ SSIM↑ LPIPS↓ | 34.59 0.9284 0.0776 | 34.40 0.9257 0.0807 | 33.42 0.9120 0.0933 | 30.65 0.8449 0.1203 | 30.60 0.8410 0.1238 | 29.91 0.8185 0.1366 | 29.31 0.8087 0.1238 | 29.29 0.8053 0.1271 | 28.78 0.7820 0.1371 | 28.74 0.8631 0.0966 | 28.48 0.8560 0.1019 | 27.65 0.8326 0.1165 |

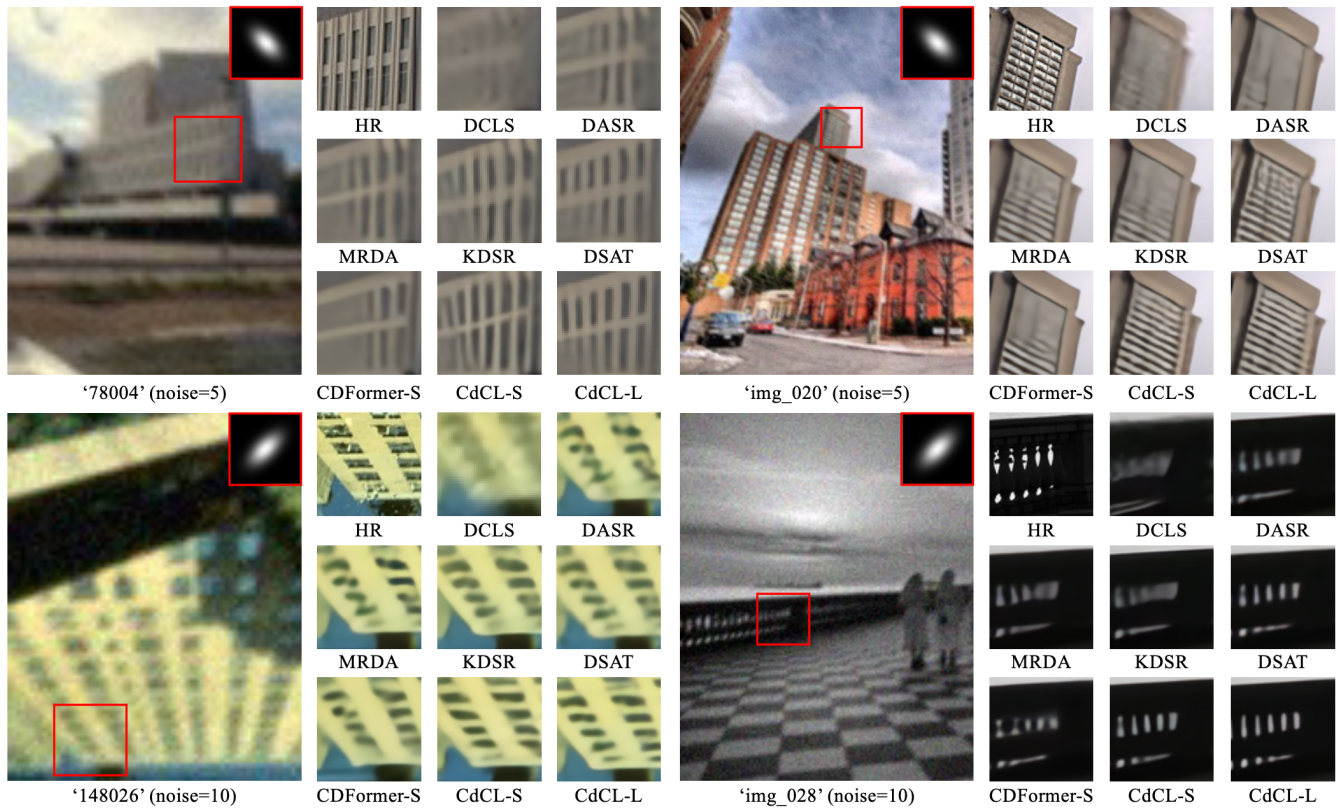

Fig. 9. ×4 SR results of "78004", "148026" on B100 and "$img\_020$", "$img\_028$" on Urban100 with anisotropic gaussian blur and noise.

settings, with the average PSNR only 0.03 dB lower than that of DCLS. Overall, our implicit estimation based method significantly reduces model complexity compared to explicit estimation based methods, thus having more advantages in practice. (2) Compared with the existing contrastive learning-based blind SR methods DASR [11] and IDMBSR [15], our method achieves significant improvement under multiple blur kernel widths. For example, on the Set5 [43], with blur kernel width of 2.4, our CdCL-S improves by 0.48 dB and 0.45 dB compared to DASR and IDMBSR, respectively, while the number of parameters is only 50% and 69% of theirs, respectively. This show the superiority of our content-decoupled contrastive learning framework. (3) Compared with the MRDA [12] and KDSR [33] which adopts meta-learning and knowledge distillation to extract the degradation representation, with blur kernel width 1.2, our CdCL-S surpasses

<const>JOURNAL OF LATEX CLASS FILES, VOL. 14, NO. 8, AUGUST 2021                                                                                                               10</const>

TABLE III
×4 SR QUANTITATIVE COMPARISON OF PSNR METRIC ON B100 UNDER DEGRADATION SETTING 2.

| Method | Param | Noise |  |  |  |  |  |  |  |  |  |
|---|---|---|---|---|---|---|---|---|---|---|---|
| *Comparison of Explicit Degradation Blind SR Methods* | | | | | | | | | | | |
| DnCNN+IKC | 650K+5.32M | 5<br>10 | 26.21<br>25.65 | 26.24<br>25.68 | 25.02<br>24.71 | 24.80<br>24.53 | 24.77<br>24.49 | 24.64<br>24.36 | 24.53<br>24.28 | 24.18<br>24.00 | 23.77<br>23.63 |
| DnCNN+DAN | 650K+4.33M | 5<br>10 | 26.18<br>25.69 | 26.14<br>25.63 | 24.89<br>24.62 | 24.64<br>24.42 | 24.61<br>24.39 | 24.52<br>24.29 | 24.37<br>24.18 | 24.01<br>23.89 | 23.63<br>23.54 |
| DnCNN+DCLS | 650K+13.63M | 5<br>10 | 26.15<br>25.67 | 26.12<br>25.62 | 24.89<br>24.62 | 24.66<br>24.43 | 24.62<br>24.40 | 24.52<br>24.29 | 24.39<br>24.19 | 24.03<br>23.91 | 23.65<br>23.56 |
| CdCL-S | 2.87M | 5<br>10 | 26.79<br>26.16 | 26.74<br>26.08 | 26.04<br>25.39 | 25.88<br>25.24 | 25.83<br>25.21 | 25.77<br>25.16 | 25.66<br>25.05 | 25.36<br>24.80 | 24.88<br>24.42 |
| *Comparison of CNN-based Implicit Degradation Blind SR Methods* | | | | | | | | | | | |
| DASR | 5.84M | 5<br>10 | 26.60<br>26.02 | 26.52<br>25.94 | 25.87<br>25.27 | 25.69<br>25.11 | 25.64<br>25.08 | 25.60<br>25.03 | 25.48<br>24.93 | 25.23<br>24.69 | 24.85<br>24.33 |
| MRDA | 5.84M | 5<br>10 | 26.73<br>26.12 | 26.68<br>26.05 | 25.98<br>25.36 | 25.80<br>25.20 | 25.77<br>25.18 | 25.71<br>25.12 | 25.59<br>25.01 | 25.29<br>24.76 | 24.82<br>24.36 |
| KDSR | 5.80M | 5<br>10 | 26.73<br>26.11 | 26.70<br>26.05 | 26.02<br>25.38 | 25.87<br>25.24 | 25.83<br>25.21 | 25.76<br>25.15 | 25.65<br>25.05 | 25.37<br>24.81 | 24.94<br>24.46 |
| CdCL-S | 2.87M | 5<br>10 | 26.79<br>26.16 | 26.74<br>26.08 | 26.04<br>25.39 | 25.88<br>25.24 | 25.83<br>25.21 | 25.77<br>25.16 | 25.66<br>25.05 | 25.36<br>24.80 | 24.88<br>24.42 |
| *Comparison of Transformer-based Implicit Degradation Blind SR Methods* | | | | | | | | | | | |
| DSAT | 15.64M | 5<br>10 | 26.71<br>26.07 | 26.66<br>26.04 | 25.95<br>25.29 | 25.80<br>25.18 | 25.74<br>25.14 | 25.66<br>25.09 | 25.55<br>24.97 | 25.30<br>24.72 | 24.85<br>24.36 |
| CDFormer-S | 11.86M | 5<br>10 | 26.77<br>26.14 | 26.69<br>26.06 | 26.00<br>25.37 | 25.85<br>25.22 | 25.82<br>25.20 | 25.75<br>25.15 | 25.64<br>25.04 | 25.34<br>24.78 | 24.97<br>24.45 |
| CdCL-L | 5.91M | 5<br>10 | 26.82<br>26.18 | 26.78<br>26.11 | 26.07<br>25.42 | 25.92<br>25.28 | 25.87<br>25.25 | 25.82<br>25.20 | 25.70<br>25.09 | 25.38<br>24.85 | 24.95<br>24.54 |

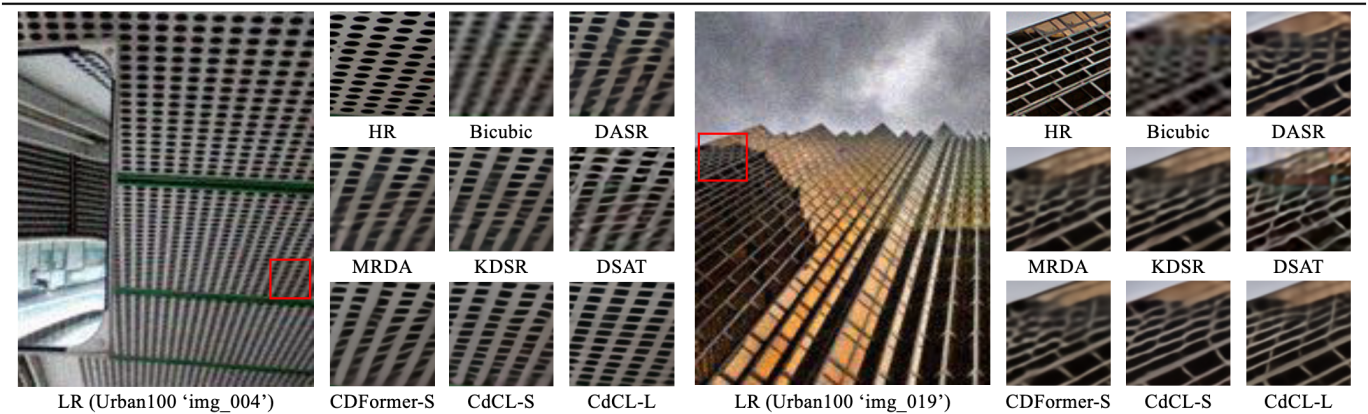

Fig. 10. The "$img\_004$", "$img\_019$" on Urban100 at ×4 SR with kernel width 1.0, noise level 10 and JPEG compression quality factor 80.

the MRDA by 0.12 dB and the KDSR by 0.13 dB (PSNR metric) on Set14, respectively. And the number of parameters and computational complexity are only about half of these methods (e.g., only 50% and 53% of MRDA, respectively). It demonstrates that the contrastive learning-based implicit blind SR paradigm still has a higher upper bound than the meta learning- and the knowledge distillation-based scheme. (4) Compared with Transformer-based DSAT and CDFormer-S, our CdCL-L still achieves SOTA performance across all degradation scenarios and evaluation metrics with less than half of their complexity, indicating that CNN architecture still has great potential in the field of implicit blind super-resolution. For scaling factor 3, the quantitative results are shown in Table II, with the LPIPS metric added. Similarly, for all datasets and different blur kernel widths, our CdCL-S and CdCL-L respectively achieve optimal results in comparison with CNN-based and Transformer-based methods, showing that our method has good robustness for different scaling.

**Qualitative evaluation.** The super-resolved results of different models on scaling factors 4 and 3 are shown in Fig. 6 and Fig. 7, respectively. The SR images generated by our method are superior to all competitors in terms of structure and low-level details. Two conclusions can be drawn: 1) compared with the explicit strategy (e.g., IKC, DAN, DCLS), implicit strategy can more comprehensively model the degradation characteristics contained in images; 2) it is feasible to construct a lightweight SR tool to accurately restore low-level details such as local structures and textures by improving the discriminability of degradation representations and the efficiency of degradation adaptation.

*2) Experiments in Degradation Setting 2*

**Quality of degradation representation.** We generate LR images from B100 benchmark [45] under *Degradation Setting 2* with scaling factor 4 and feed them to Basemodel, DASR [11], MRDA [12], KDSR [33], CDFormer-S [36] and CdCL-S to generate corresponding degradation representations. Subsequently, the t-SNE plots are shown in Fig. 8. In this challenging degradation scenarios, MRDA, KDSR, and our method still effectively distinguish different degradations and cluster representations of the same degradation, whereas the degradation space bulit by DASR and CDFormer-S is of poor quality. This result highlights the critical importance of a well-designed degradation representation learning framework.

TABLE IV
×4 SR QUANTITATIVE COMPARISON OF PSNR METRIC UNDER DEGRADATION SETTING 3.

| Dataset | Method | Param(M) | bic | b2.0 | n20 | j60 | b2.0n20 | b2.0j60 | n20j60 | b2.0n20j60 | Average |
|---|---|---|---|---|---|---|---|---|---|---|---|
| Set14 | Bicubic | — | 25.00 | 25.34 | 21.77 | 24.29 | 21.91 | 24.51 | 21.46 | 21.73 | 23.25 |
| | *Comparison of CNN-based Implicit Degradation Blind SR Methods* ||||||||||
| | DASR | 5.84 | 28.29 | 27.77 | 26.16 | 26.16 | 25.11 | 25.22 | 25.33 | 24.55 | 26.07 |
| | MRDA | 5.84 | 28.55 | 28.31 | 26.25 | 26.37 | 25.18 | 25.49 | 25.53 | 24.57 | 26.28 |
| | KDSR | 5.80 | 28.62 | 28.36 | 26.28 | 26.42 | 25.22 | 25.56 | 25.55 | 24.67 | 26.33 |
| | CdCL-S | 2.87 | 28.69 | 28.38 | 26.34 | 26.47 | 25.20 | 25.52 | 25.61 | 24.64 | 26.35 |
| | *Comparison of Transformer-based Implicit Degradation Blind SR Methods* ||||||||||
| | DSAT | 15.64 | 28.90 | 28.86 | 25.72 | 26.10 | 25.24 | 25.34 | 25.18 | 24.18 | 26.19 |
| | CDFormer-S | 11.86 | 28.71 | 28.62 | 26.31 | 26.44 | 25.31 | 25.67 | 25.56 | 24.73 | 26.41 |
| | CdCL-L | 5.91 | 28.91 | 28.89 | 26.44 | 26.59 | 25.41 | 25.76 | 25.68 | 24.80 | 26.56 |
| B100 | Bicubic | — | 24.63 | 25.40 | 21.56 | 24.06 | 21.90 | 24.65 | 21.22 | 21.72 | 23.14 |
| | *Comparison of CNN-based Implicit Degradation Blind SR Methods* ||||||||||
| | DASR | 5.84 | 27.43 | 27.21 | 25.58 | 25.70 | 24.80 | 25.13 | 24.97 | 24.41 | 25.65 |
| | MRDA | 5.84 | 27.55 | 27.42 | 25.63 | 25.80 | 24.84 | 25.20 | 25.05 | 24.46 | 25.74 |
| | KDSR | 5.80 | 27.57 | 27.49 | 25.64 | 25.82 | 24.89 | 25.27 | 25.06 | 24.48 | 25.77 |
| | CdCL-S | 2.87 | 27.63 | 27.46 | 25.66 | 25.85 | 24.89 | 25.27 | 25.08 | 24.48 | 25.79 |
| | *Comparison of Transformer-based Implicit Degradation Blind SR Methods* ||||||||||
| | DSAT | 15.64 | 27.77 | 27.72 | 25.26 | 25.47 | 24.46 | 24.99 | 24.57 | 24.13 | 25.54 |
| | CDFormer-S | 11.86 | 27.64 | 27.58 | 25.65 | 25.83 | 24.89 | 25.32 | 25.06 | 24.50 | 25.80 |
| | CdCL-L | 5.91 | 27.78 | 27.76 | 25.71 | 25.90 | 24.96 | 25.37 | 25.11 | 24.54 | 25.89 |
| Urban100 | Bicubic | — | 21.89 | 22.54 | 20.00 | 21.50 | 20.36 | 22.02 | 19.74 | 20.20 | 21.03 |
| | *Comparison of CNN-based Implicit Degradation Blind SR Methods* ||||||||||
| | DASR | 5.84 | 25.59 | 24.76 | 23.96 | 23.81 | 22.79 | 22.79 | 23.22 | 22.28 | 23.65 |
| | MRDA | 5.84 | 25.97 | 25.57 | 24.12 | 24.05 | 23.03 | 23.09 | 23.43 | 22.50 | 23.97 |
| | KDSR | 5.80 | 26.04 | 25.64 | 24.19 | 24.14 | 23.09 | 23.25 | 23.49 | 22.54 | 24.04 |
| | CdCL-S | 2.87 | 26.17 | 25.75 | 24.26 | 24.20 | 23.15 | 23.23 | 23.56 | 22.56 | 24.11 |
| | *Comparison of Transformer-based Implicit Degradation Blind SR Methods* ||||||||||
| | DSAT | 15.64 | 26.68 | 26.36 | 24.12 | 23.81 | 22.91 | 22.89 | 22.79 | 22.00 | 23.94 |
| | CDFormer-S | 11.86 | 26.23 | 25.87 | 24.26 | 24.20 | 23.17 | 23.32 | 23.54 | 22.63 | 24.15 |
| | CdCL-L | 5.91 | 26.70 | 26.39 | 24.55 | 24.45 | 23.40 | 23.54 | 23.77 | 22.82 | 24.45 |

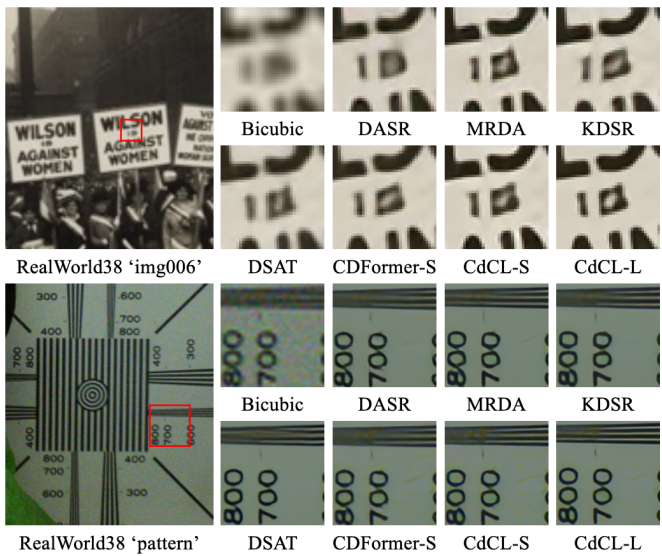

Fig. 11. ×4 SR results of the "$img006$" and "$pattern$" on RealWorld38 [48] under degradation setting 2.

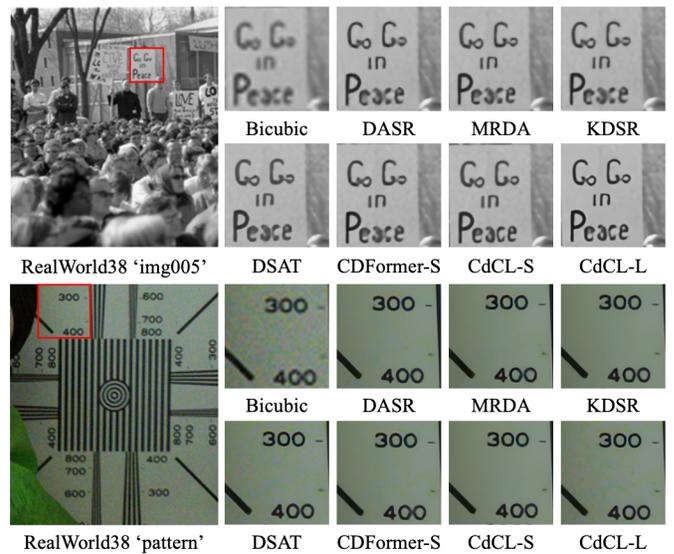

Fig. 12. ×4 SR results of the real image "$img005$" and "$pattern$" on RealWorld38 [48] under degradation setting 3.

**Quantitative evaluation.** We select 9 different anisotropic blur kernels and add 2 additional noise levels (i.e., 5, 10). For IKC [9], DAN [10] and DCLS [28] models without denoising ability, we use DnCNN [53] to denoise LR images before SR processing for a fair comparison. Table III shows the PSNR results of our method with other competitors on the B100 benchmark with more complex textures. It can be seen that: 1) our method is significantly superior to explicit estimation based methods in terms of performance and complexity. For example, compared to the SOTA DCLS [28] method under this paradigm, the average PSNR of CdCL-S is higher than DCLS by over 0.93 dB, once again indicating the superiority of implicit degradation estimation. 2) under the implicit estimation paradigm, our method achieves the best results in almost all degradation settings (i.e. various noise levels and blur kernels), whether compared to CNN-based methods or Transformer-based methods, and with half the parameters of competitors. For example, CdCL-S outperforms KDSR [33] by over 0.01 dB for the vast majority of noise levels and blur kernel conditions, but the parameters are less than half of KDSR. Furthermore, compared to Transformer-based methods [35], [36] with larger number of parameters, our CNN-based approach once again demonstrates absolute advantages, with an average PSNR of 0.05 dB higher than CDFormer-S under various degradation conditions.

TABLE V
EFFECT OF THE CdIDM AND DaIDAM ON PSNR PERFORMANCE OF CdCL FOR ALL DEGRADATION SETTINGS.

| | | Setting1 | | | | | | | | |
|---|---|---|---|---|---|---|---|---|---|---|
| | | Set14 | | | B100 | | | Urban100 | | |
| | Param(M) | 1.2 | 2.4 | 3.6 | 1.2 | 2.4 | 3.6 | 1.2 | 2.4 | 3.6 |
| DASR | 5.84 | 28.44 | 28.28 | 27.45 | 27.52 | 27.43 | 26.83 | 25.69 | 25.44 | 24.66 |
| DASR-A | 5.84 | 28.53↑ | 28.30↑ | 27.47↑ | 27.60↑ | 27.49↑ | 26.89↑ | 25.78↑ | 25.54↑ | 24.75↑ |
| DASR-B | 4.09 | 28.64↑ | 28.46↑ | 27.49↑ | 27.62↑ | 27.55↑ | 26.93↑ | 26.04↑ | 25.78↑ | 24.87↑ |
| CdCL | 2.87 | 28.79 | 28.64 | 27.73 | 27.73 | 27.63 | 27.00 | 26.36 | 26.06 | 25.20 |
| CdCL-A | 2.87 | 28.64↓ | 28.50↓ | 27.63↓ | 27.65↓ | 27.58↓ | 26.95↓ | 25.81↓ | 25.86↓ | 24.99↓ |
| CdCL-B | 4.39 | 28.57↓ | 28.48↓ | 27.54↓ | 27.61↓ | 27.56↓ | 26.95↓ | 25.90↓ | 25.63↓ | 25.03↓ |
| | | Setting2 (Set14, noise=5) | | | | | | | | |
| | Param(M) | [kernel] | [kernel] | [kernel] | [kernel] | [kernel] | [kernel] | [kernel] | [kernel] | [kernel] |
| DASR | 5.84 | 27.25 | 27.18 | 26.37 | 26.16 | 26.09 | 25.96 | 25.85 | 25.52 | 25.04 |
| DASR-A | 5.84 | 27.32↑ | 27.39↑ | 26.40↑ | 26.23↑ | 26.13↑ | 26.02↑ | 25.93↑ | 25.59↑ | 25.07↑ |
| DASR-B | 4.09 | 27.44↑ | 27.36↑ | 26.49↑ | 26.32↑ | 26.24↑ | 26.12↑ | 26.02↑ | 25.67↑ | 25.09↑ |
| CdCL | 2.87 | 27.68 | 27.60 | 26.66 | 26.45 | 26.41 | 26.35 | 26.23 | 25.77 | 25.13 |
| CdCL-A | 2.87 | 27.51↓ | 27.45↓ | 26.58↓ | 26.40↓ | 26.34↓ | 26.20↓ | 26.08↓ | 25.76↓ | 25.11↓ |
| CdCL-B | 4.39 | 27.45↓ | 27.38↓ | 26.32↓ | 26.15↓ | 26.13↓ | 25.90↓ | 25.92↓ | 25.58↓ | 25.04↓ |
| | | Setting3 (Urban100) | | | | | | | | |
| | Param(M) | bic | b2.0 | n20 | j60 | b2.0n20 | b2.0j60 | n20j60 | b2.0n20j60 | Average |
| DASR | 5.84 | 25.59 | 24.76 | 23.96 | 23.81 | 22.79 | 22.79 | 23.22 | 22.28 | 23.65 |
| DASR-A | 5.84 | 25.56↓ | 25.17↑ | 23.96 | 23.90↑ | 22.81↑ | 22.84↑ | 23.24↑ | 22.26↓ | 23.72↑ |
| DASR-B | 4.09 | 25.78↑ | 25.50↑ | 24.15↑ | 24.08↑ | 23.05↑ | 23.20↑ | 23.43↑ | 22.50↑ | 23.96↑ |
| CdCL | 2.87 | 26.17 | 25.75 | 24.26 | 24.20 | 23.15 | 23.23 | 23.56 | 22.56 | 24.11 |
| CdCL-A | 2.87 | 25.98↓ | 25.71↓ | 24.15↓ | 23.98↓ | 23.11↓ | 23.20↓ | 23.45↓ | 22.31↓ | 23.99↓ |
| CdCL-B | 4.39 | 26.13↓ | 25.59↓ | 24.23↓ | 24.16↓ | 23.07↓ | 23.05↓ | 23.52↓ | 22.47↓ | 24.03↓ |

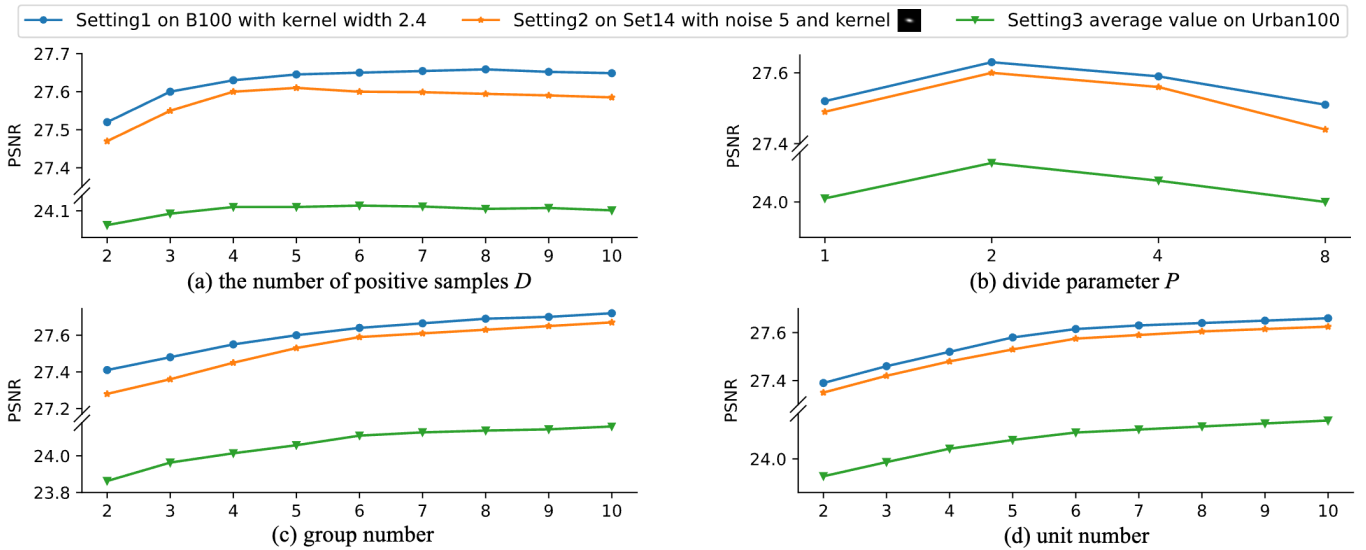

Fig. 13. The ablation study of different hyperparameters. a) Effect of the number of positive samples $D$ in cycle shift sampling. b) Effect of the divide parameter $P$ in CdIDM. c) Effect of the number of DAGs in DaIDAM. d) Effect of the number of DaDAUs in DAGs.

**Qualitative evaluation.** Fig. 9 presents the visual results of each method under *Degradation Setting 2*. Our method recovers richer, clearer texture details even under strong noise. In contrast, the result of DCLS [28] is very blurry and lack texture; the SR images generated by DASR [11], MRDA [12], KDSR [33], DSAT [35] and CDFormer-S [36] partially restore texture details, but they are not accurate.

*3) Experiments in Degradation Setting 3*

To further verify the generalization of our method for a wider range of degradation scenarios, we retrain DASR [11],

TABLE VI
EFFECT OF DIFFERENT DADAU STRUCTURE ON PSNR PERFORMANCE OF CDCL FOR ALL DEGRADATION SETTINGS.

| Setting1 | | | | | | | | | |
|---|---|---|---|---|---|---|---|---|---|
| | Set14 | | | B100 | | | Urban100 | | |
| | 1.2 | 2.4 | 3.6 | 1.2 | 2.4 | 3.6 | 1.2 | 2.4 | 3.6 |
| DASR | 28.44 | 28.28 | 27.45 | 27.52 | 27.43 | 26.83 | 25.69 | 25.44 | 24.66 |
| CdCL+S | 28.65 | 28.42 | 27.61 | 27.66 | 27.56 | 26.91 | 26.06 | 25.75 | 24.92 |
| CdCL+C | 28.64 | 28.46 | 27.60 | 27.63 | 27.54 | 26.92 | 26.09 | 25.81 | 25.00 |
| CdCL+$C_{ns}$ | 28.69 | 28.55 | 27.70 | 27.66 | 27.57 | 26.96 | 26.18 | 25.89 | 25.10 |
| CdCL+$C_s$ | 28.79 | 28.64 | 27.73 | 27.73 | 27.63 | 27.00 | 26.36 | 26.06 | 25.20 |
| Setting2 (Set14, noise=5) | | | | | | | | | |
| | 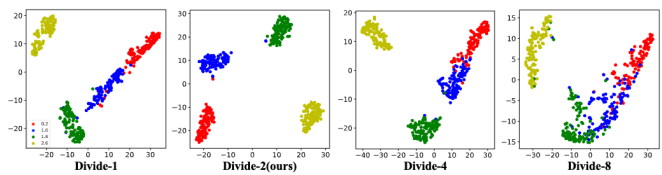 | | | | | | | | |
| DASR | 27.25 | 27.18 | 26.37 | 26.16 | 26.09 | 25.96 | 25.85 | 25.52 | 25.04 |
| CdCL+S | 27.45 | 27.38 | 26.51 | 26.34 | 26.31 | 26.21 | 26.08 | 25.73 | 25.07 |
| CdCL+C | 27.52 | 27.43 | 26.52 | 26.31 | 26.21 | 26.10 | 25.91 | 25.47 | 24.70 |
| CdCL+$C_{ns}$ | 27.54 | 27.44 | 26.54 | 26.35 | 26.33 | 26.19 | 26.03 | 25.65 | 25.12 |
| CdCL+$C_s$ | 27.68 | 27.60 | 26.66 | 26.45 | 26.41 | 26.35 | 26.23 | 25.77 | 25.13 |
| Setting3 (Urban100) | | | | | | | | | |
| | bic | b2.0 | n20 | j60 | b2.0n20 | b2.0j60 | n20j60 | b2.0n20j60 | Average |
| DASR | 25.59 | 24.76 | 23.96 | 23.81 | 22.79 | 22.79 | 23.22 | 22.28 | 23.65 |
| CdCL+S | 25.93 | 25.54 | 24.12 | 24.03 | 22.98 | 23.14 | 23.21 | 22.38 | 23.92 |
| CdCL+C | 26.08 | 25.75 | 24.21 | 24.08 | 23.10 | 23.15 | 23.31 | 22.40 | 24.01 |
| CdCL+$C_{ns}$ | 26.12 | 25.81 | 24.22 | 24.12 | 23.13 | 23.21 | 23.42 | 22.43 | 24.06 |
| CdCL+$C_s$ | 26.17 | 25.75 | 24.26 | 24.20 | 23.15 | 23.23 | 23.56 | 22.56 | 24.11 |

Fig. 14. The t-SNE [17] plots of degradation representation distributions for different divide parameter $P$. 4 different isotropic blur kernels are selected under *Degradation Setting 1* on B100.

MRDA [12], KDSR [33], DSAT [35], CDFormer-S [36], CdCL-S and CdCL-L under *Degradation Setting 3*, and evaluate them on the Set14, B100, and Urban100 datasets using 8 degradation combinations described in reference [47], i.e., {bic, b2.0, n20, j60, b2.0n20, b2.0j60, n20j60, b2.0n20j60}. The comparison results are shown in Table IV. It can be seen that no matter in what kind of degradation scenarios, our method can obtain obvious performance advantages, proving that our method has better generalization and robustness. Qualitative results are displayed in Fig. 10, where the degradation parameters of LR image are blur kernel width 1.0, noise level 10, and JPEG compression quality factor 80. As can be observed from the highlighted regions, our method generates sharper texture details and smoother edges.

*4) Experiments in real degradation*

Finally, we validate the generalization of our method and other competitors for more challenging real degradation, where the degradation parameters are complex and unknown. Since there is no corresponding HR image, we only show the qualitative comparisons between different methods. Under the condition of *Degradation Setting 2* with scaling factor 4, the restored results of "$img006$" and "$pattern$" from RealWorld38 [48] by different models are shown in the Fig. 11. It can be seen that even for degradation-unknown real images containing textual and digital parts, our method can better recover a lot of high-frequency details in these LR images than other competitors. In addition, we show the reconstruction results of the LR image "$img005$" and "$pattern$" from RealWorld38 [48] by different models under the *Degradation Setting 3*, as shown in Fig. 12. In this challenging case, our method successfully restores clearer and sharper parts of letters and numbers, demonstrating its distinct advantage in recovering fine texture details.

### C. Ablation Study

This section provides an in-depth analysis of our method, including the effects of the proposed Content-decoupled Implicit Degradation Modeling (CdIDM) technique and Detail-aware Implicit Degradation Adaptation Module (DaIDAM), as well as the impact of different hyperparameter settings on performance. We perform full ablation experiments on all three degradation settings based on the CdCL small-version.

*1) Effect of the CdIDM and DaIDAM*

To verify the contribution of CdIDM and DaIDAM in the contrastive learning-based blind SR network, we perform component swapping between our CdCL and DASR [11].

**For implicit degradation modeling.** We swap the degradation representation learning components between CdCL and DASR, resulting in two variants, CdCL-A and DASR-A. CdCL-A represents replacing its CdIDM with the degradation modeling framework of DASR, while keeping the rest of the network unchanged. DASR-A represents using our CdIDM to train DASR's degradation estimator. The experimental results are shown in Table V. The SR performance of DASR-A has significantly improved, with an average PSNR 0.07 dB higher than DASR on all datasets and different degradation settings. At the same time, the performance of CdCL-A has significantly decreased compared to CdCL, with an average decrease of 0.12 dB in PSNR. The above results demonstrate that the proposed implicit degradation modeling technique: 1) can extract more precise degradation representations; 2) has good compatibility and can be seamlessly integrated into contrastive learning-based implicit blind SR methods.

TABLE VII
EFFECT OF LR FEATURES INTEGRATION OF DADAU ON PSNR PERFORMANCE FOR ALL DEGRADATION SETTINGS.

| | | Setting1 | | | | | | | | |
|---|---|---|---|---|---|---|---|---|---|---|
| | | Set14 | | | B100 | | | Urban100 | | |
| channel+LR | spatial+LR | 1.2 | 2.4 | 3.6 | 1.2 | 2.4 | 3.6 | 1.2 | 2.4 | 3.6 |
| ✓ | ✓ | 28.79 | 28.62 | 27.74 | 27.76 | 27.63 | 27.01 | 26.33 | 26.05 | 25.18 |
| | | 28.73 | 28.59 | 27.68 | 27.65 | 27.60 | 26.96 | 26.27 | 25.95 | 25.16 |
| | ✓ | 28.72 | 28.56 | 27.69 | 27.67 | 27.59 | 26.96 | 26.21 | 25.93 | 25.11 |
| ✓ | | 28.79 | 28.64 | 27.73 | 27.73 | 27.63 | 27.00 | 26.36 | 26.06 | 25.20 |
| | | Setting2 (Set14, noise=5) | | | | | | | | |
| channel+LR | spatial+LR |  |  |  |  |  |  |  |  |  |
| ✓ | ✓ | 27.68 | 27.58 | 26.63 | 26.46 | 26.40 | 26.35 | 26.25 | 25.80 | 25.10 |
| | | 27.62 | 27.57 | 26.62 | 26.40 | 26.35 | 26.30 | 26.21 | 25.71 | 25.09 |
| | ✓ | 27.62 | 27.55 | 26.58 | 26.41 | 26.35 | 26.28 | 26.23 | 25.77 | 25.07 |
| ✓ | | 27.68 | 27.60 | 26.66 | 26.45 | 26.41 | 26.35 | 26.23 | 25.77 | 25.13 |
| | | Setting3 (Urban100) | | | | | | | | |
| channel+LR | spatial+LR | bic | b2.0 | n20 | j60 | b2.0n20 | b2.0j60 | n20j60 | b2.0n20j60 | Average |
| ✓ | ✓ | 26.15 | 25.72 | 24.25 | 24.21 | 23.13 | 23.23 | 23.56 | 22.59 | 24.10 |
| | | 26.14 | 25.70 | 24.23 | 24.16 | 23.11 | 23.20 | 23.51 | 22.52 | 24.07 |
| | ✓ | 26.10 | 25.68 | 24.20 | 24.16 | 23.10 | 23.21 | 23.49 | 22.54 | 24.06 |
| ✓ | | 26.17 | 25.75 | 24.26 | 24.20 | 23.15 | 23.23 | 23.56 | 22.56 | 24.11 |

**For implicit degradation adaptation.** We swap the degradation adaptation modules between CdCL and DASR, resulting in two variants, CdCL-B and DASR-B. The ×4 SR experimental results are presented in Table V. It can be observed that the use of DaIDAM can bring significant performance improvements and reduce the number of model parameters, showing that it can more effectively adapt implicit degradation representations to specific LR image features. For example, DASR-B surpasses DASR with an average PSNR 0.21 dB for all datasets and degradation settings. In contrast, the average PSNR of CdCL-B is 0.18 dB lower than that of CdCL.

*2) Effect of different CdIDM settings*

The proposed core component CdIDM includes two key operations: cyclic shift sampling and divide-combine.

**Cyclic shift sampling operation** is used to implement data augmentation in the preprocessing stage, in which parameter $D$ ($D <$ batchsize) determines the number of augmented positive samples. To verify the impact of $D$ on SR performance, we gradually increase $D$ from 2 to 10 and the results are displayed in Fig. 13(a). In theory, having more positive samples is more advantageous for learning discriminative degradation representations. However, in actual evaluation, we find that the model performance steadily improved as $D$ increased from 2 to 4. But as $D$ continues to increase, the model performance reaches saturation and even begin to gradually decline, which may be due to the redundant features caused by limited data based excessive augmenting damaging the robustness of the model to degradation distribution shifts [54].

**Divide-combine operation** is also used to augment the sample diversity of each degradation category, where the parameter $P$ determines the scale of augmentation (dividing the input patch into $P^2$ sub-patches on average). We set a series of increasing $P$-values to verify its impact on SR performance, as shown in Fig. 13(b). When $P = 1$ (i.e., no divide-combine operation), the PSNR has a significant decrease over all degradation settings, proving the effectiveness of this operation. When $P \geq 4$, the PSNR gradually decreases. This is because as $P$ increases, the size of the sub-patch seen by the estimator decreases, making it difficult to obtain sufficient information from sub-patches to learn degradation representations. To further investigate this, we conduct t-SNE visualizations of the representations extracted by the degradation estimator under different $P$ values. As shown in Fig. 14, these results highlight that $P = 2$ is the optimal setting for balancing sampling diversity and image information integrity. This setting ensures that the degradation representations are both discriminative and robust, providing the best foundation for SR recovery. Note that the divide-combine operation is only used during training, so it does not increase the inference burden.

*3) Effect of different DaIDAM settings*

The key to the effectiveness of DaIDAM lies in the continuous adaptation between implicit degradation representation and LR image features, which involves two hyperparameter settings: the number of DaDAUs in the DAG and the number of DAGs in the DaIDAM. Specifically, we first fix the number of DaDAUs contained in each DAG to be 6, and gradually increase the number of DAGs in the DaIDAM from 2 to 10. Subsequently, we fix the number of DAGs in the DaIDAM to be 6, and gradually increase the number of DaDAUs in each DAG from 2 to 10. As can be seen in Fig. 13(c), (d), as the number of DAGs or DaDAUs increases, the SR performance steadily improves for different degradation settings, proving the importance of continuous adaptation. But as the number grows from 6 to 10, the performance improvement gradually slows down and the training time becomes longer accordingly. To balance the model complexity and performance, we choose the number of DAGs in the DaIDAM and the number of DaDAUs in the DAG to be 6 and 6, respectively.

*4) Effect of different DaDAU structure on SR performance*

DaDAU consists of two branches that modulate degradation representations from channel and spatial perspectives and fuse them with LR image features. Following the Degradation Setting 1 with scaling factor 4, we built 4 models according to Sec. III, namely CdCL+S, CdCL+C, CdCL+$C_s$ and CdCL+$C_{ns}$. Among them, CdCL+S means that only the spatial adaptation branch is added in DaDAU. CdCL+C means that only the channel adaptation branch is added in DaDAU. CdCL+$C_s$ means that in the channel adaptation branch, the two FC layers are shared between channel degradation representation $D_{channel}$ and pooled LR image features $GAP(F_{LR})$. CdCL+$C_{ns}$ means that the two FC layers of the channel adaptation branch are not shared between $D_{channel}$ and $GAP(F_{LR})$. Table. VI shows the comparisons. It can be seen that even if only the spatial branch or the channel branch are used, the performance of our method is significantly better than DASR, which proves the superiority of the proposed adaptation branches. In addition, the FC layers sharing mechanism in the channel adaptation branch achieves better results, which we believe is due to the fact that considering LR features in the degradation representation modulation process enhances DaDAU's awareness of specific image characteristics, thereby improving the degradation adaptation effect.

To further explore the effectiveness of integrating LR image features into the modulation process, we investigate the impact of different LR features integration settings on SR performance, as shown in Table VII. Specifically, channel+LR denotes the integration of LR image features into the channel adaptation branch, while spatial+LR refers to their integration into the spatial adaptation branch. The ablation results show that introducing image features into the channel adaptation branch significantly improves SR performance across all degradation settings. In contrast, integrating LR features into the spatial adaptation branch provides limited performance improvement, and even has adverse effects in some degradation scenarios. This means that in the spatially-consistent degradation problem, the modulation of the degradation representation does not need to overly focus on the local details of the image.

## V. CONCLUSION

This paper focuses on the blind image SR task, and propose a new Content-decoupled Contrastive Learning-based blind image super-resolution (CdCL) framework, which includes two key innovations: Content-decoupled Implicit Degradation Modeling (CdIDM) technique and Detail-aware Implicit Degradation Adaptation Module (DaIDAM). The former innovatively introduces a cyclic shift sampling strategy to ensure that degradation representation learning is not interfered by task-independent information at the data level, thereby improving the purity and discriminability of the learned degradation space. The latter improves the fusion effect between degradation representations and LR image features by enhancing the basic adaptation unit's perception of image details, thereby achieving high-quality adaptation and significantly reducing model complexity. Extensive comparative experiments show that our method achieves highly competitive quantitative and qualitative results in the blind SR task with very low computational costs, and sufficient ablation studies demonstrate the effectiveness of the proposed components. Our work reveals an important insight that enhancing the discriminability of degradation representations and improving adaptation network's perception of image details is a more parameter-efficient path for developing blind SR tools.